\documentclass[10pt]{article}

\usepackage{arxiv}

\usepackage[utf8]{inputenc} 
\usepackage[T1]{fontenc}    
\usepackage{hyperref}       
\usepackage{url}            
\usepackage{booktabs}       
\usepackage{amsfonts}       
\usepackage{nicefrac}       
\usepackage{microtype}      
\usepackage{mathtools}
\usepackage{lipsum}
\usepackage{mathpazo}
\usepackage{amsmath}
\usepackage[frozencache]{minted}
\usepackage{tikz}
\usepackage{caption}
\usepackage{subcaption}
\usepackage{graphicx}
\usetikzlibrary{matrix,chains,positioning,decorations.pathreplacing,arrows}
\usetikzlibrary{positioning}

\usepackage[style=phys, backend=biber]{biblatex}
\usepackage{titlesec}
\titlespacing*{\section}
{0pt}{3.5ex plus 1ex minus .2ex}{1.3ex plus .2ex}
\titlespacing*{\subsection}
{0pt}{2.0ex plus 1ex minus .2ex}{0.3ex plus .2ex}

\DeclareMathOperator*{\argmax}{arg\,max}
\DeclareMathOperator*{\argmin}{arg\,min}

\DeclarePairedDelimiterX{\infdivx}[2]{(}{)}{%
  #1\;\delimsize\|\;#2%
}
\newcommand{\kldist}{\mathbb{KL}\infdivx}

\title{Exploring Variational Deep Q Networks}

\author{
  A.H.~Bell-Thomas \\
  Computer Laboratory \\
  University of Cambridge \\
  \texttt{Alexander.Bell-Thomas@cl.cam.ac.uk}
}

\begin{document}
\maketitle

\begin{abstract}
\footnotesize{
This study provides a research-ready implementation of Tang and Kucukelbir's Variational Deep Q Network, a novel approach to maximising the efficiency of exploration in complex learning environments using Variational Bayesian Inference, using the Edward PPL. Alongside reference implementations of both Traditional and Double Deep Q Networks, a small novel contribution is presented --- the Double Variational Deep Q Network, which incorporates improvements to increase the stability and robustness of inference-based learning. Finally, an evaluation and discussion of the effectiveness of these approaches is discussed in the wider context of Bayesian Deep Learning.}
\end{abstract}

\section*{Introduction}
\paragraph{} The balance between state-space exploration and exploiting known effective actions is one of the fundamental problems in learning against complex environments. A promising approach to tackling this is by building an approximation of the potential value of exploration --- this can be achieved by using Bayesian methods on agents' uncertainty about the optimality of the current policy. This concept was first introduced over 20 years ago by Dearden et. al.,~\cite{10.5555/295240.295801} and has been explored in numerous works since. The focus of this study shall be the application of Variational Bayesian Inference~\cite{Blei_2017} in this domain, primarily inspired by the work of Tang and Kucukelbir.~\cite{tang2017variational}

\paragraph{} This paper accompanies an open-source reference implementation\footnote{\url{https://github.com/HarriBellThomas/VDQN}} of all four algorithms discussed, one of which is a novel contribution. Further, it extends the evaluation and discussion contributed by the original authors, including commentary on concepts discussed in works published since. 

\section{Background}

\paragraph{} Based on Markov Decision Processes (MDPs), Reinforcement Learning (RL) is one of the most fundamental modern machine learning techniques. RL is, at its core, a stochastic optimisation approach for unknown MDPs --- it observes the state of the environment and takes action according to the current policy it holds. The goal is simple: maximise the reward gained, as defined by the model's reward function, by choosing the best action from any given state.

\vspace{-4mm}
\begin{align}
\label{eqn:q}
Q(s,a) : S \times A \rightarrow R
\end{align}

\clearpage
\paragraph{} An optimal policy (\ref{eqn:rl}) is defined as a policy maximising the discounted cumulative reward from any state.

\vspace{-4mm}
\begin{align}
\label{eqn:rl}
    \pi^{*} &= \argmax_{\pi} \mathbb{E}_{\sigma\sim\pi(\cdot | s_0)} \Big[ \: \sum_{t \in \sigma} Q^{\pi}(s_t, a_t) \cdot \gamma^{t} \: \Big]
\end{align}

\paragraph{} Where $\gamma$ is the discount constant, $0 \leq \gamma \leq 1$, and $\sigma$ is the sequence of actions and states chosen by the policy, $\pi$; this defines a single \textit{episode}.

\vspace{6mm}
\subsection*{Q Learning}
\paragraph{} Q Learning is a \textit{model-free} RL approach introduced by Watkins~\cite{watkinsqlearning}. In additional to learning $\pi^{*}$, it concurrently iteratively learns an optimal reward function, $Q^{*}$, for the problem domain (\ref{eqn:ql}). $\alpha$ is the learning rate, $0 < \alpha \leq 1$.

\vspace{-4mm}
\begin{align}
\label{eqn:ql}
Q^{next}(s_t,a_t) \leftarrow (1-\alpha) * Q(s_t,a_t) + \alpha * (r_t + \gamma * \argmax_{a} Q(s_{t+1},a))
\end{align}

\paragraph{} An error function is required to guide both the policy and reward functions to optimal forms --- for this Bellman's equations are used.~\cite{bellman1954} The Bellman error is a measure of the decay in expectation of future advantage as an agent executes an action under policy $\pi$ (\ref{eqn:be}). Conceptually, $J(\pi)$ measures the expected loss in reward earning-potential as the agent progresses through an episode. It follows that a pure optimal policy, $\pi^{*}$, will conserve earning potential (have zero Bellman error) across agents' actions (\ref{eqn:be-opt}).

\vspace{-3mm}
\begin{align}
\label{eqn:be}
J(\pi) = \mathbb{E}_{a_t\sim\pi(\cdot | s_t)}  \Big[ \big( Q^{\pi}(s_{t},a_{t}) - \argmax_a \mathbb{E}[r_{t} + \gamma \cdot Q^{\pi}(s_{t+1}, a)] \big)^{2} \Big]
\end{align}

\vspace{-5mm}
\begin{align}
\label{eqn:be-opt}
Q^{*}(s_t, a_t) = \argmax_{a} \mathbb{E} \Big[ r_t + \gamma \cdot Q^{*}(s_{t+1}, a) \Big]
\end{align}

\vspace{6mm}
\subsection*{Deep Q Networks (DQN)}

\paragraph{} Deep RL extends traditional RL methods by leveraging a neural network (NN) to approximate the policy $\pi$. This neural policy, $\pi_{\theta}$, where $\theta$ represents the parameter vector of the policy NN, can be updated directly using differentiation (\ref{eqn:dqn-params}).

\vspace{-6mm}
\begin{align}
\label{eqn:dqn-params}
    \theta &\leftarrow \theta + \alpha \cdot \nabla_{\theta} \cdot J(\theta)
\end{align}

\paragraph{} In practicality, the loss function is discretised into $N$ samples. Additionally, as introduced by Mnih et. al. in the seminal DQN paper,~\cite{mnih2013playing} it is common to maintain two networks concurrently --- and active network, $Q^{\theta}$, and a target network, $Q^{\theta^{-}}$. Both networks have identical architectures, but the target is updated more slowly than the active one to minimise perturbations in performance; this is effectively a debouncing method. This gives (\ref{eqn:dqn-be-a}), the approximate Bellman error for a DQN.

\vspace{-3mm}
\begin{align}
\label{eqn:dqn-be-a}
J(\theta) \approx \frac{1}{N} \sum_{j=1}^{N} ( Q^{\theta}(s_{j},a_{j}) - \argmax_{a'} ( r_{t} + \gamma \cdot Q^{\theta^{-}}(s_{j}', a') ))^2
\end{align}

\newpage
\subsection*{Variational Deep Q Networks (VDQN)}

\paragraph{} Introduced by Tang and Kucukelbir,~\cite{tang2017variational} the Variational DQN exploits an equivalence between a surrogate objective (\ref{eqn:vdqn-obj}) and variational inference loss.\footnote{David Blei has published an excellent introduction briefing on the underlying mathematical concepts of variational inference.~\cite{blei-vi}} Taking interaction with the environment as a generative model, VDQNs construct prior and posterior distributions on $\theta$, the network's parameters, with respect to recorded observations, $D$.

\vspace{-5mm}
\begin{align}
\text{Prior} & \rightarrow p(\theta) \\
\text{Posterior} & \rightarrow p(\theta | D) = \frac{p(\theta, D)}{p(D)}
\end{align}
\vspace{2mm}

\paragraph{} As explained in~\cite{blei-vi}, calculating the posterior is generally extremely challenging. The core concept underpinning variational methods is the selection of a family of distributions over the
latent variables with its own variational parameters; here $q_{\phi}(\theta)$ is used to approximate the posterior. This has now become an optimisation problem over a class of tractable distributions, $q$, parameterised by $\phi$, in order to find the one most similar to $p(\theta | D)$. This is iteratively solved using gradient descent on the Kullback-Leibler divergence between approximate and true posterior distributions (\ref{eqn:vdqn-kl}), and is suitable for use as a surrogate posterior distribution.~\cite{Blei_2017,kucukelbir2016automatic,ranganath2013black}

\vspace{-5mm}
\begin{align}
\label{eqn:vdqn-kl}
\phi \leftarrow \min\limits_{\phi} \kldist{q_{\phi}(\theta)}{p(\theta | D)}
\end{align}

\paragraph{} A valuable insight contributed by Tang and Kucukelbir is that in order to encourage efficient exploration $q_{\phi}(\theta)$ must have a sufficiently large entropy. Thus the goal of a VDQN is to find $\phi$ that minimises Bellman loss across a highly dispersed set of candidate policies. With $\lambda > 0$, a regularisation constant, a VDQN's objective function is expressed as follows.

\vspace{-3mm}
\begin{align}
\label{eqn:vdqn-obj}
\mathbb{E}_{\theta\sim q_{\phi}(\theta)} \Big[ \big( Q^{\theta}(s_{j},a_{j}) - \max\limits_{a'}\mathbb{E}[r_{j} + \gamma Q^{\theta}({s}_{j}', a')] \big)^{2} \Big] - \lambda\mathbb{H}(q_{\phi}(\theta))
\end{align}

\paragraph{} The debouncing method from DQNs is implemented in a similar way here: the active network is parameterised by $\theta\sim q_{\phi}(\theta)$ and the target network by $\theta^{-}\sim q_{\phi^{-}}(\theta^{-})$.

\subsection*{Double Q Learning}
\paragraph{} Q-Learning occasionally performs poorly in some stochastic environments; this is ascribed to large overestimation of action rewards in training. To remedy this, van Hasselt proposed Double Q Learning:~\cite{NIPS2010_3964, hasselt2015deep} there are two different approaches (one per paper) --- this study uses the latter.~\cite{hasselt2015deep} The concept is relatively simple --- to discourage action overestimation incorporate the target network as an infrequently changing frame of reference. To achieve this the $Q$ function update operation (previously defined in (\ref{eqn:ql})) is changed as follows. Note that the $\argmax$ function is over the target network instead of the active one.

\vspace{-4mm}
\begin{align}
\label{eqn:ddqn}
Q^{next}(s_t,a_t) \leftarrow (1-\alpha) * Q(s_t,a_t) + \alpha * (r_t + \gamma * \argmax_{a} Q'(s_{t+1},a))
\end{align}

\clearpage

\section{Prototype}
\paragraph{} The open-source implementation that accompanies this report implements four Q learning implementations; DQN, Double DQN, VDQN, and Double VDQN. Double VDQN (DVDQN) is a novel extension of the original VDQN design, adapting Mnih et. al.'s stabilisation techniques to improve observed volatility of the variational inference approach.

\subsection*{Double Variational Deep Q Networks}
\begin{figure}
    \centering
    \hfill%
    \begin{subfigure}{.74\textwidth}
      \centering
      \includegraphics[width=\hsize]{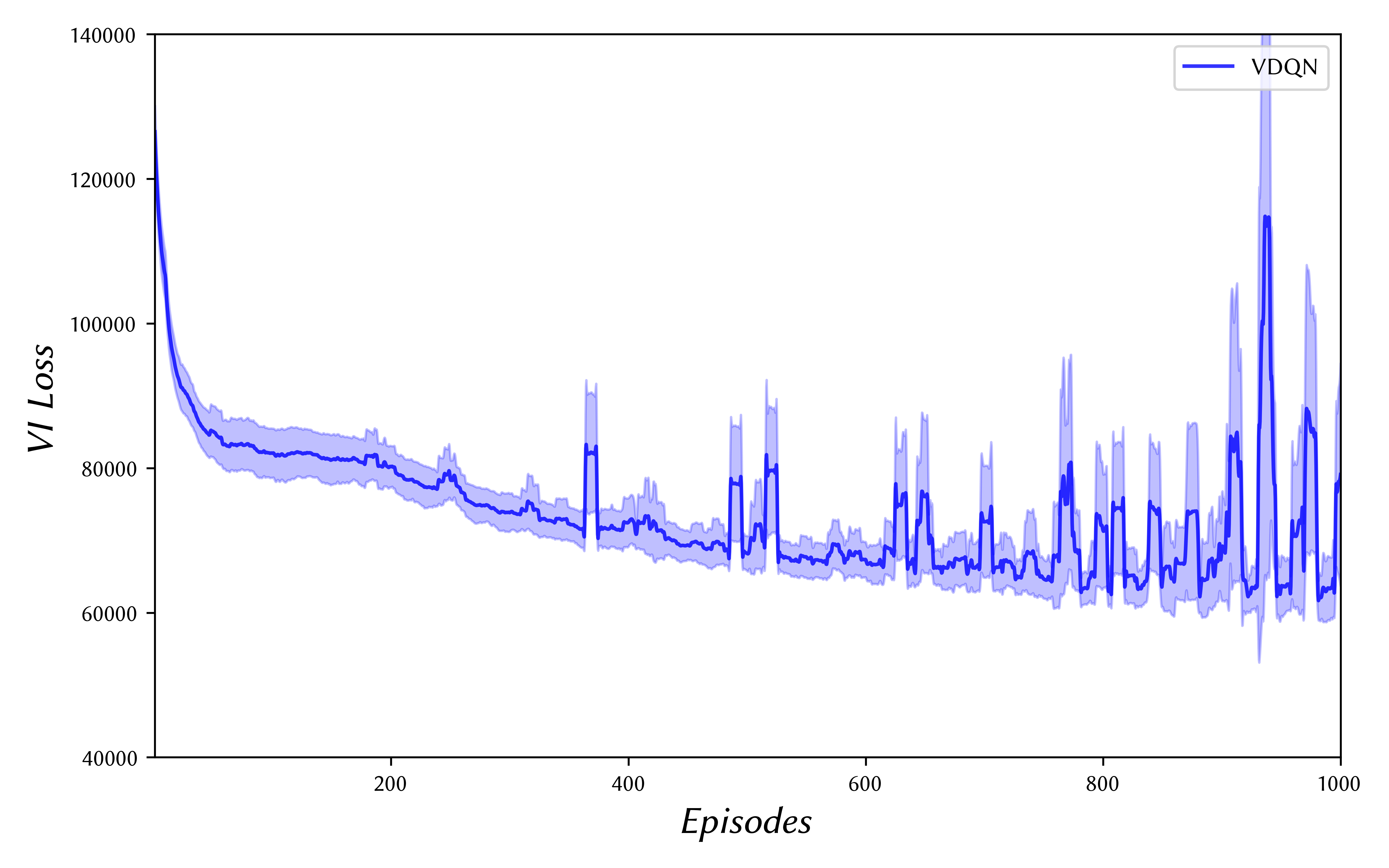}
    \end{subfigure}%
    \hspace{0.13\textwidth}
    \caption{\texttt{MountainCar-v0} --- VDQN  VI loss during training.}
    \label{fig:mountaincar-vi-loss}
\end{figure}

\paragraph{} As the Evaluation section will show in detail, the results seen from the VDQN are highly promising. They clearly show that the new approach does appear to provide highly efficient state-space exploration, as claimed. Figure~\ref{fig:mountaincar-vi-loss} is a plot of observed VI loss\footnote{VI loss is the delta between the prior and approximate posterior distributions after training; lower is better.} whilst training the model in the OpenAI Gym's \texttt{MountainCar-v0} environment; the VDQN's perturbations are caused by the inference model 'jumping' to correct itself. This behaviour is reminiscent of the erraticism DDQN aimed to tackle. 

\paragraph{} The hypothesis held when this study commenced was that the root cause of this erratic behaviour is the introduction of the $-\lambda \mathbb{H}(q_{\phi}(\theta))$ bias term. It, just as seen with the vanilla DQN's $\argmax$ term, is a guiding approximation that could be susceptible to over-approximating its true value. Thus an approach close to that employed by DDQNs is used --- DVDQNs manage the speed of updates to $\phi$, the variational parameter, to indirectly have the same effect as (\ref{eqn:ddqn}); importantly, both the entropy of the variational family and the qualitative behaviours of the sampled set of $\theta$ parameters (and therefore the policies $\pi_{\theta}$) are influenced. This technique is implemented using a modified form of (\ref{eqn:ddqn}) to update the posterior distribution of the inference model.

\subsection*{Implementation}
\begin{figure}
    \centering
    \resizebox{0.6\linewidth}{!}{


\tikzset{%
  every neuron/.style={
    circle,
    draw,
    minimum size=1cm
  },
  neuron missing/.style={
    draw=none, 
    scale=4,
    text height=0.333cm,
    execute at begin node=\color{black}$\vdots$
  },
}

\begin{tikzpicture}[x=1.5cm, y=1.5cm, >=stealth]

\foreach \m/\l [count=\y] in {1,2,3,missing,4}
  \node [every neuron/.try, neuron \m/.try] (input-\m) at (0,2.5-\y) {};

\foreach \m [count=\y] in {1,missing,2}
  \node [every neuron/.try, neuron \m/.try ] (hidden1-\m) at (2,2-\y*1.25) {};

\foreach \m [count=\y] in {1,missing,2}
\node [every neuron/.try, neuron \m/.try ] (hidden1-\m) at (2,2-\y*1.25) {};

\foreach \m [count=\y] in {1,missing,2}
  \node [every neuron/.try, neuron \m/.try ] (hidden2-\m) at (4,2-\y*1.25) {};

\foreach \m [count=\y] in {1,missing,2}
  \node [every neuron/.try, neuron \m/.try ] (output-\m) at (6,1.5-\y) {};

\foreach \l [count=\i] in {1,2,3,x}
  \draw [<-] (input-\i) -- ++(-1,0)
    node [above, midway] {$I_\l$};

\foreach \l [count=\i] in {1,n}
  \node [above] at (hidden1-\i.north) {$H_{1,\l}$};

\foreach \l [count=\i] in {1,n}
  \node [above] at (hidden2-\i.north) {$H_{2,\l}$};

\foreach \l [count=\i] in {1,y}
  \draw [->] (output-\i) -- ++(1,0)
    node [above, midway] {$O_\l$};

\foreach \i in {1,...,4}
  \foreach \j in {1,...,2}
    \draw [->] (input-\i) -- (hidden1-\j);

\foreach \i in {1,...,2}
  \foreach \j in {1,...,2}
    \draw [->] (hidden1-\i) -- (hidden2-\j);

    \foreach \i in {1,...,2}
    \foreach \j in {1,...,2}
    \draw [->] (hidden2-\i) -- (output-\j);


\end{tikzpicture}
}
    \caption{The dual hidden layer design used by all four algorithms.}
    \label{fig:nn}
\end{figure}
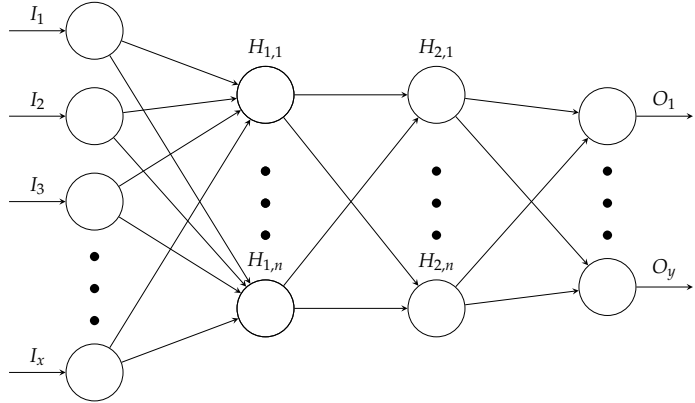

\paragraph{} All four algorithms are implemented using the same technologies in a unified Python framework; using the framework and reproducing the experiments described in this report will be discussed in Appendix \ref{appendix:reproduce}. The framework uses Tensorflow, Chainer, and the Edward probabilistic programming library. All models use the same network structure (Figure \ref{fig:nn}) --- two equally-sized fully-connected hidden layers.\footnote{The experiments performed for this report used 100 nodes per layer.} The framework currently only supports a subset of OpenAI Gym's environments due to varying state/action space representations --- \texttt{Box(x,)} state spaces and \texttt{Discrete(y)} action spaces are supported.\footnote{More information about this can be found at: \url{https://github.com/openai/gym/wiki/Table-of-environments}.}

\paragraph{DQN / DDQN Implementation} The reference non-variational algorithms were implemented primarily using the Chainer library.\footnote{\url{https://github.com/chainer/chainer}} A rectified network design is used (the activation function is build from ReLU calls), and both use the traditional $\epsilon$-greedy technique (\ref{eqn:epsilon}) to facilitate early episode exploration~\cite{sutton2011reinforcement} (with a configurable linear decay (1.0 $\rightarrow$ 0.1) period of 30 episodes).

\vspace{-4mm}
\begin{align}
\label{eqn:epsilon}
\pi(s)= 
\begin{dcases}
    \text{max}_a Q(s,a),& \text{with probability } 1 - \epsilon \\
    \text{random action},& \text{otherwise}
\end{dcases}
\end{align}

\paragraph{VDQN / DVDQN Implementation} The neural network structures for the variational models are constructed manually using a combination of TensorFlow and Edward to conform to their respective designs. Again, the neural networks used are rectified. The generative model built to represent the environment uses a Gaussian distribution centered on a greedily chosen action from the current $Q^{\theta}$ function ((\ref{eqn:vdqn-1}), $\sigma = 0.01$ by default). As suggested in the original paper, an improper uniform prior\footnote{A uniform distribution with an integral not necessarily equal to 1.} is used. The structure for the Bayesian neural network was inspired by the Edward whitepaper,~\cite{tran2016edward} which describes how to initialise all of the required components, and the original VDQN paper.

\vspace{-3mm}
\begin{align}
\label{eqn:vdqn-1}
d_i \sim \mathbb{N}(Q^{\theta}(s, a), \sigma^2)
\end{align}

\paragraph{} \textit{A note on the Edward library:} the original plan for this project had been to use the new Edward2 library\footnote{\url{https://github.com/google/edward2}} with TensorFlow v2 but there were a number of blocking problems. At the time of writing Edward2 has no stable/release version and currently lacks fully-featured inference functionality. The upgrade guide suggests building a trace update workflow manually, but given the complexity of the inference model required by this project this was deemed to be too far out of scope to be viable. Instead, Edward (v1.3.5)\footnote{\url{http://edwardlib.org/}} and the most recent version of TensorFlow it supports (v1.4.1) are used --- this version of TensorFlow is also used for the non-variational implementations to ensure fairness.

\section{Evaluation}
\begin{figure}
    \centering
    \hfill%
    \begin{subfigure}{.75\textwidth}
      \centering
      \includegraphics[width=\hsize]{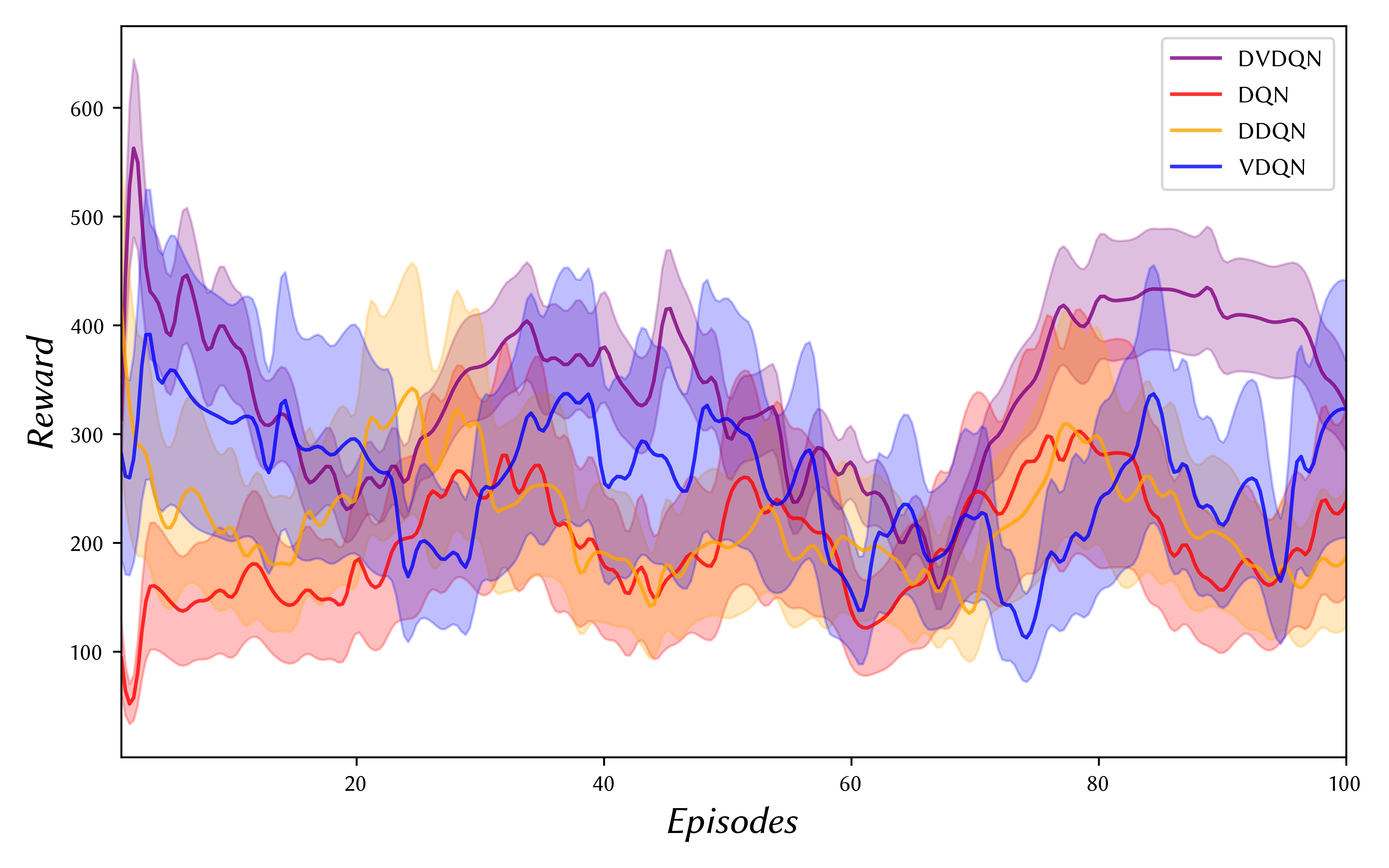}
      \vspace{-7mm}
      \caption{\texttt{SpaceInvaders-v0}}
      \label{fig:bw-bidir-1:d}
    \end{subfigure}%
    \hspace{0.125\textwidth}

    \medskip

    \hfill%
    \begin{subfigure}{.75\textwidth}
      \centering
      \includegraphics[width=\hsize]{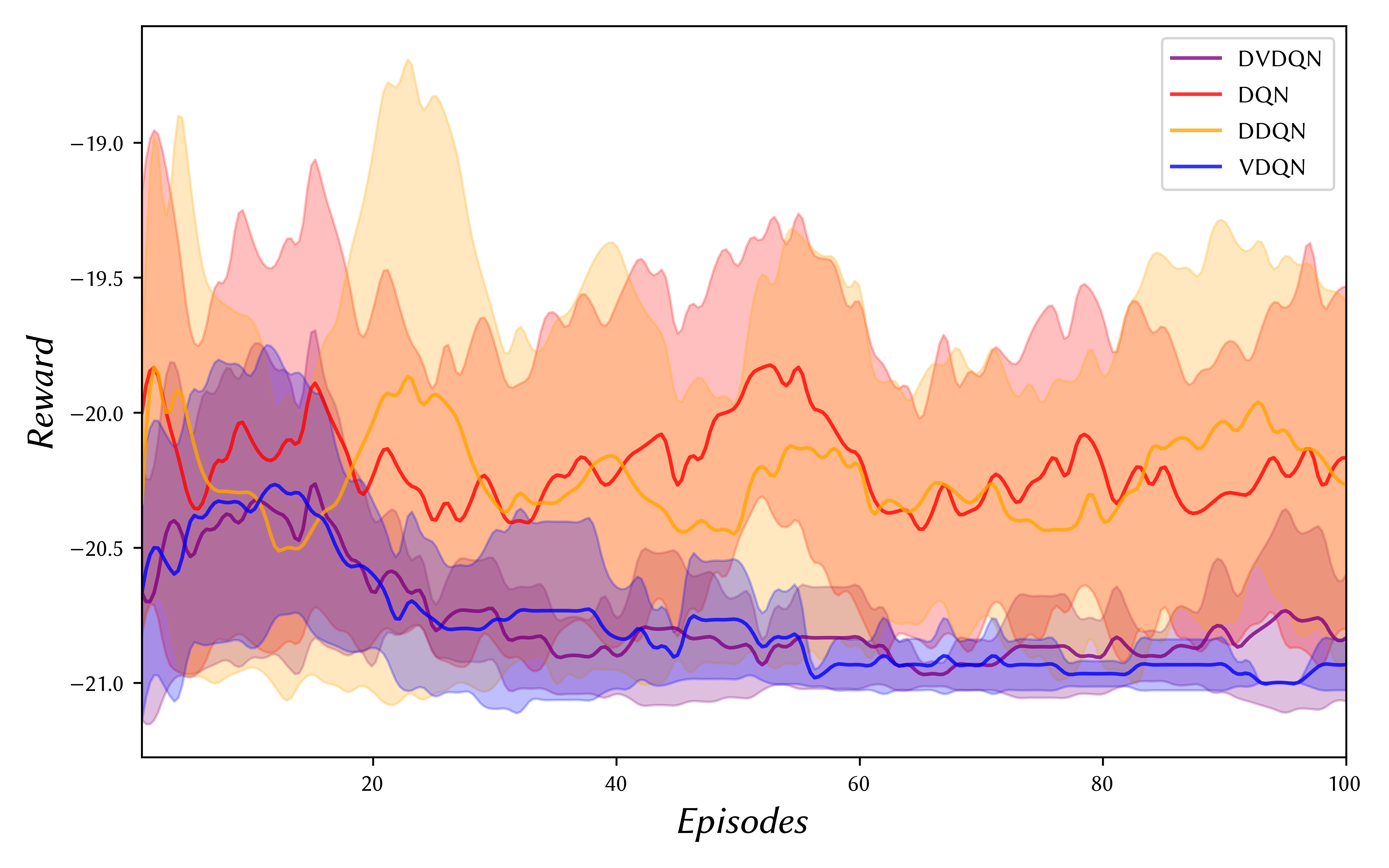}
      \vspace{-7mm}
      \caption{\texttt{Pong-v0}}
      \label{fig:bw-bidir-1:d}
    \end{subfigure}%
    \hspace{0.125\textwidth}

    \captionsetup{justification=centering}
    \caption{Training curves observed for 2 complex OpenAI Gym Atari game environments. \\ Error bounds are $\pm s$, the sample std. dev.}
\label{fig:basic-atari-rewards}
\end{figure}

\paragraph{} A set of experiments from OpenAI's Gym\footnote{\url{https://gym.openai.com/}} were used to assess the performances of the different algorithms. The original VDQN paper gave a very brief look at the performance in 4 control problems; Figure \ref{fig:basic-rewards} recreates this, and Figures \ref{fig:basic-atari-rewards} and \ref{fig:basic-vi-bellmans} extends it. For each algorithm in each experiment the highest performing parameter set has been plotted. All graphs have been placed at the end of the section for clarity.

\subsection{Discussion of Results}

\begin{figure}
\centering
\begin{subfigure}{.49\textwidth}
  \centering
  \includegraphics[width=\hsize]{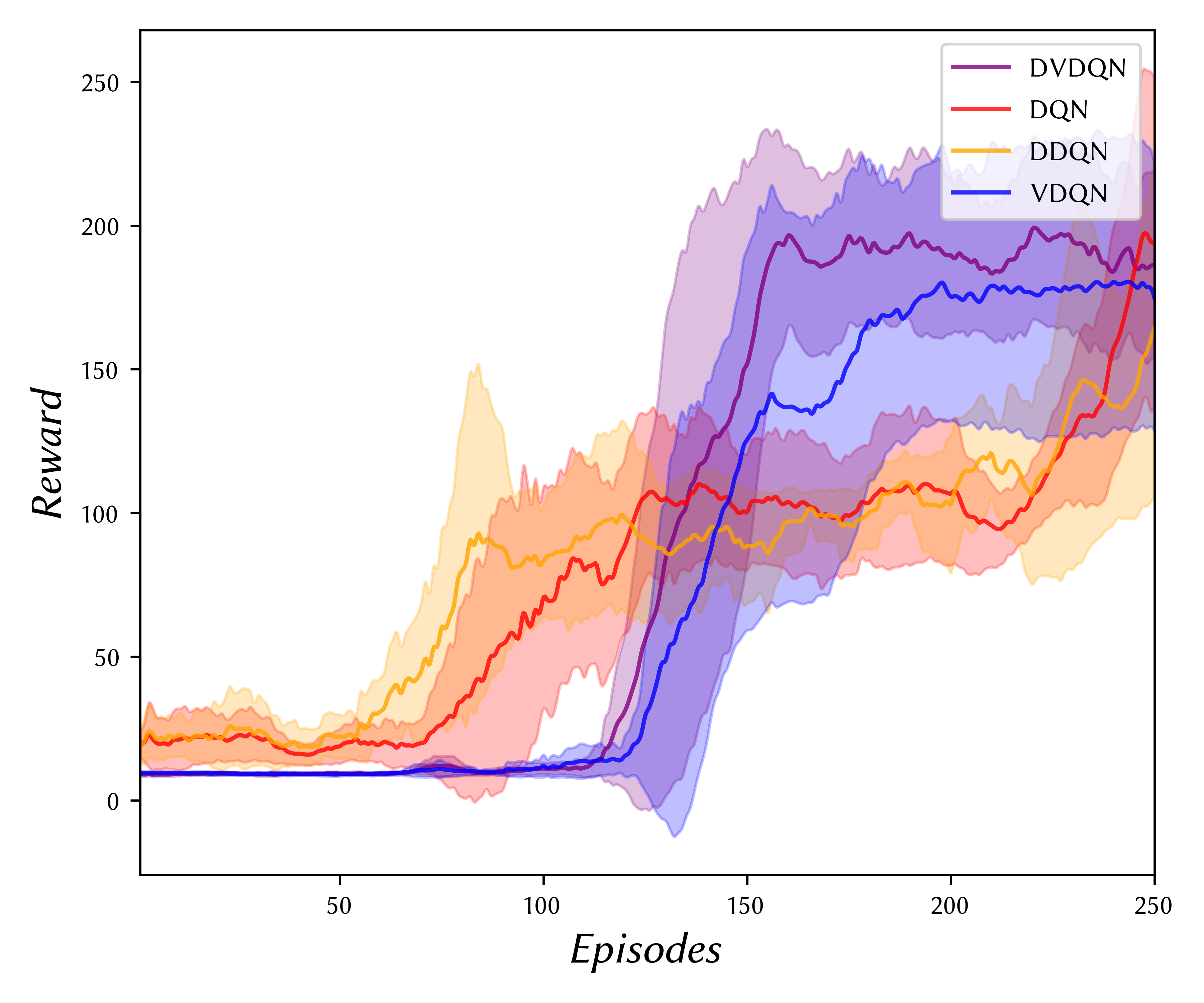}
  \vspace{-7mm}
  \caption{\texttt{CartPole-v0}}
  \label{fig:bw-bidir-1:a}
\end{subfigure}%
\hfill%
\begin{subfigure}{.49\textwidth}
  \centering
  \includegraphics[width=\hsize]{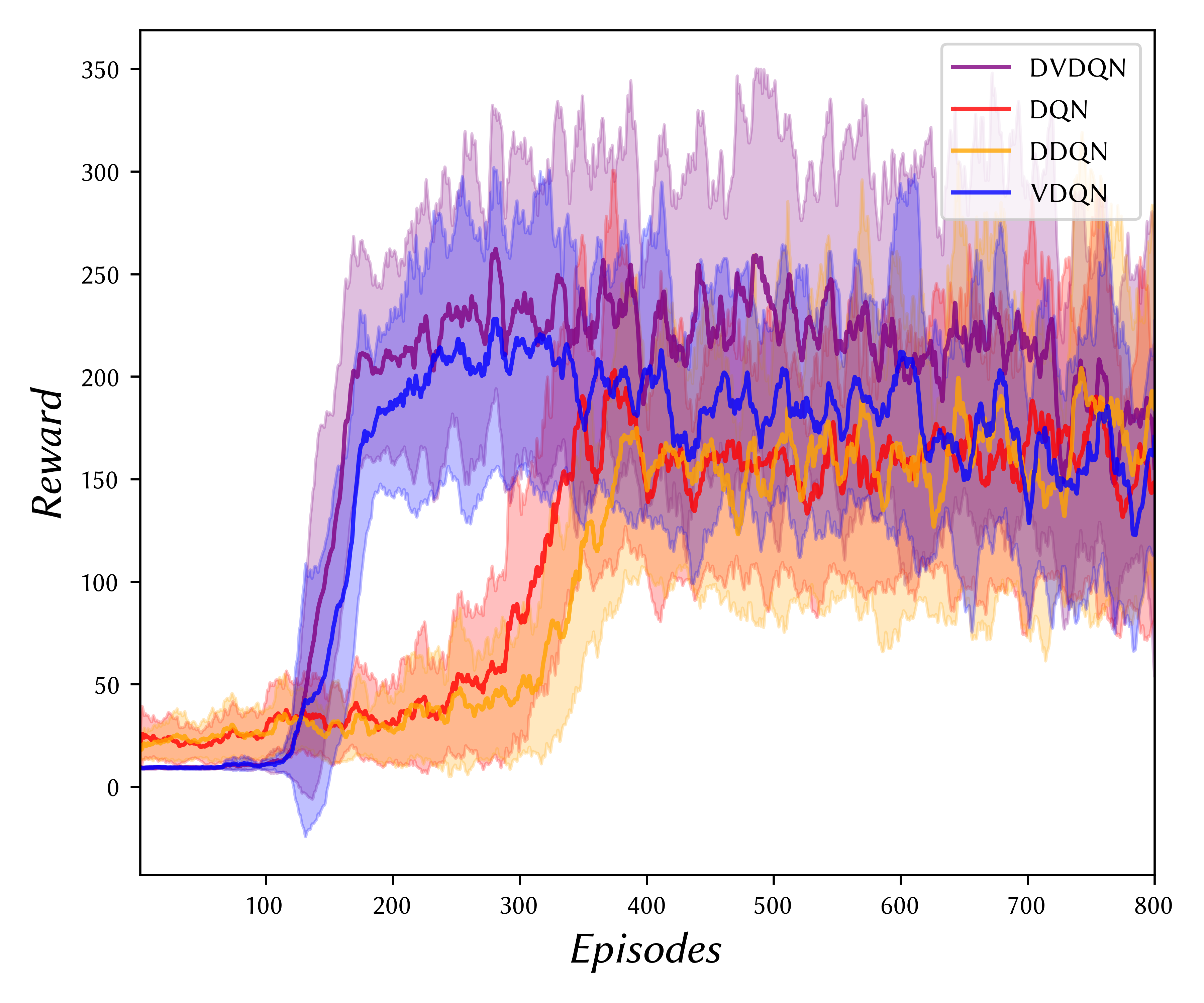}
  \vspace{-7mm}
  \caption{\texttt{CartPole-v1}}
  \label{fig:bw-bidir-1:b}
\end{subfigure}%

\medskip

\clearpage

\begin{subfigure}{.49\textwidth}
  \centering
  \includegraphics[width=\hsize]{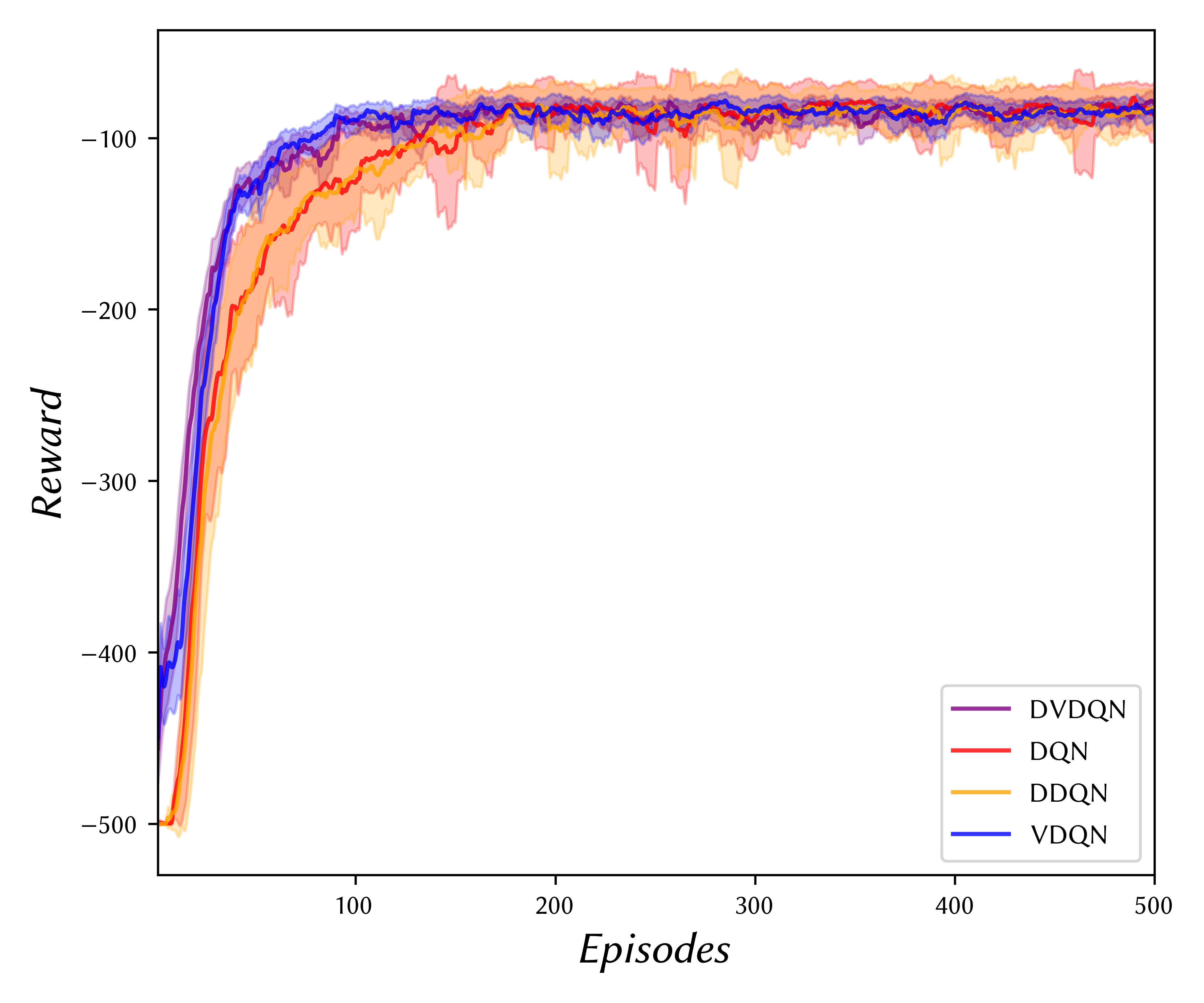}
  \vspace{-7mm}
  \caption{\texttt{Acrobot-v1}}
  \label{fig:bw-bidir-1:a}
\end{subfigure}%
\hfill%
\begin{subfigure}{.49\textwidth}
  \centering
  \includegraphics[width=\hsize]{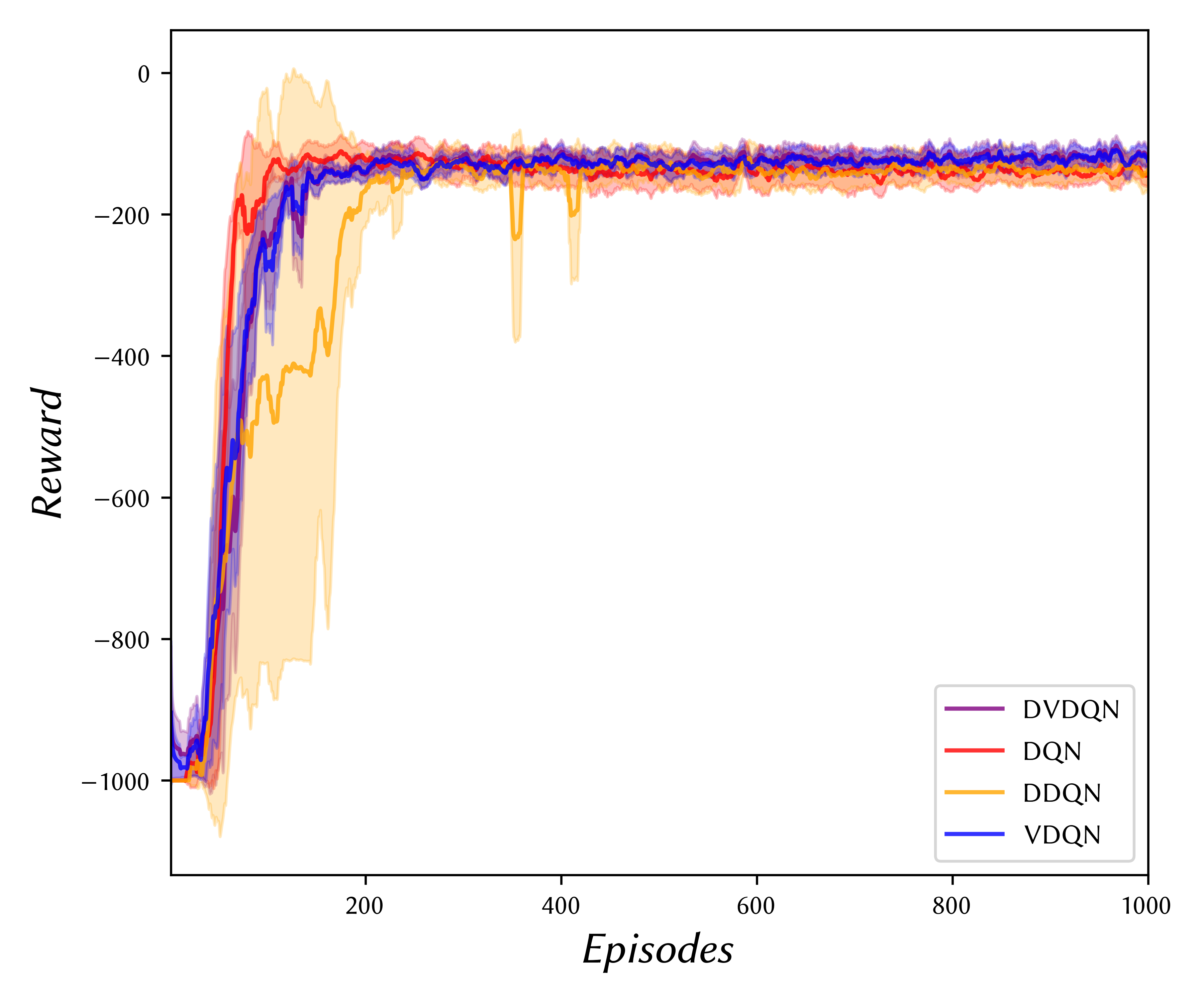}
  \vspace{-7mm}
  \caption{\texttt{MountainCar-v0}}
  \label{fig:bw-bidir-1:b}
\end{subfigure}%

\medskip

\captionsetup{justification=centering}
\caption{Rewards observed during training for all four algorithms. \\ Error bounds are $\pm s$, the sample std. dev.}
\label{fig:basic-rewards}
\end{figure}

\paragraph{} On simpler tasks (Figure \ref{fig:basic-rewards}), such as the \texttt{CartPole} experiments, the variational approaches make progress much faster, but, as clear in \texttt{CartPole-v1}, they take longer to converge on the optimal policy. As Tang and Kucukelbir point out, this is likely caused by the hindrance of the entropy bonus term as it encourages exploration even when a suspected optimal policy has been found.

\begin{figure}
\centering
\begin{subfigure}{.49\textwidth}
  \centering
  \includegraphics[width=\hsize]{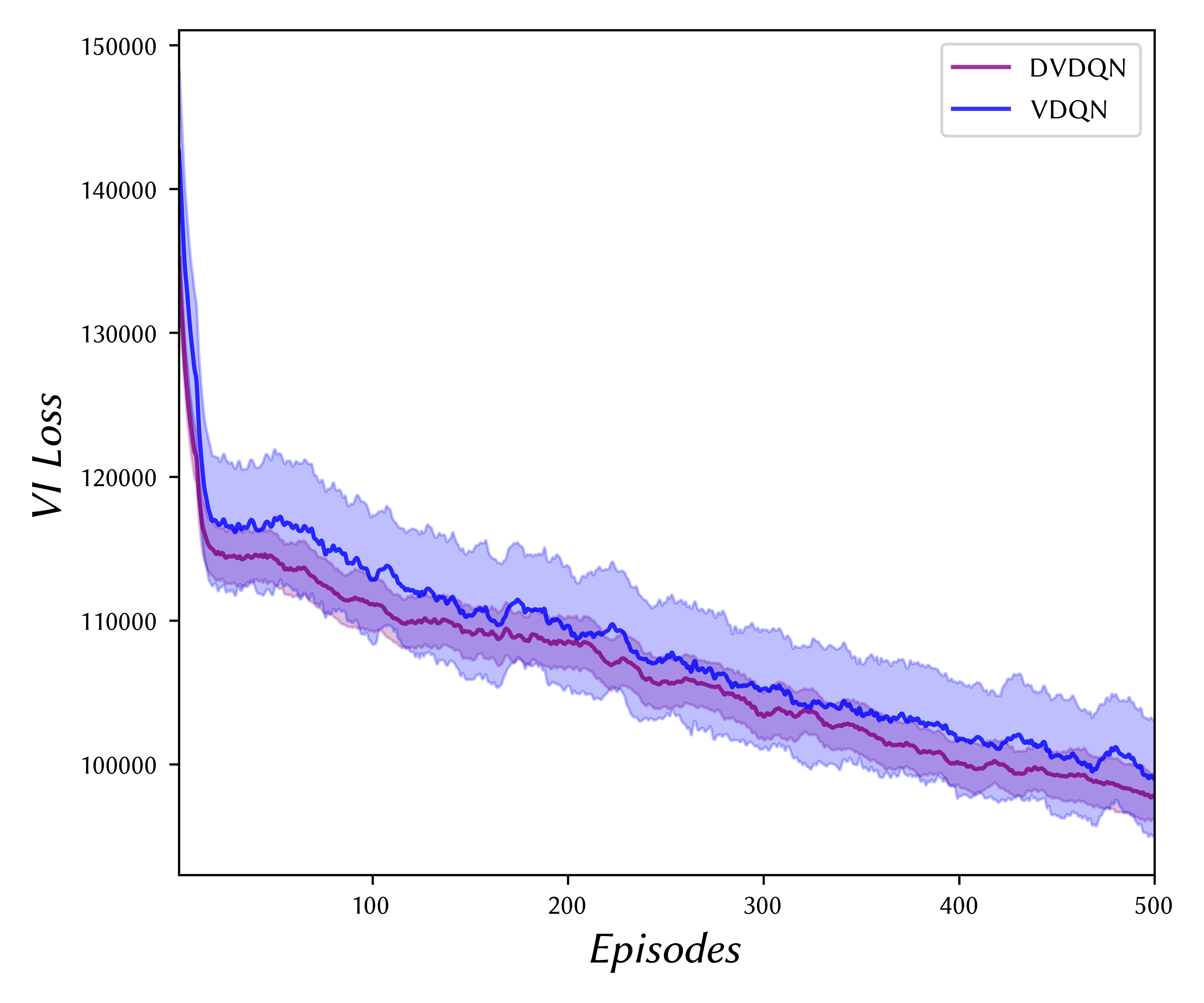}
  \vspace{-7mm}
  \caption{\texttt{Acrobot-v1} --- VI Loss}
  \label{fig:bw-bidir-1:a}
\end{subfigure}%
\hfill%
\begin{subfigure}{.49\textwidth}
  \centering
  \includegraphics[width=\hsize]{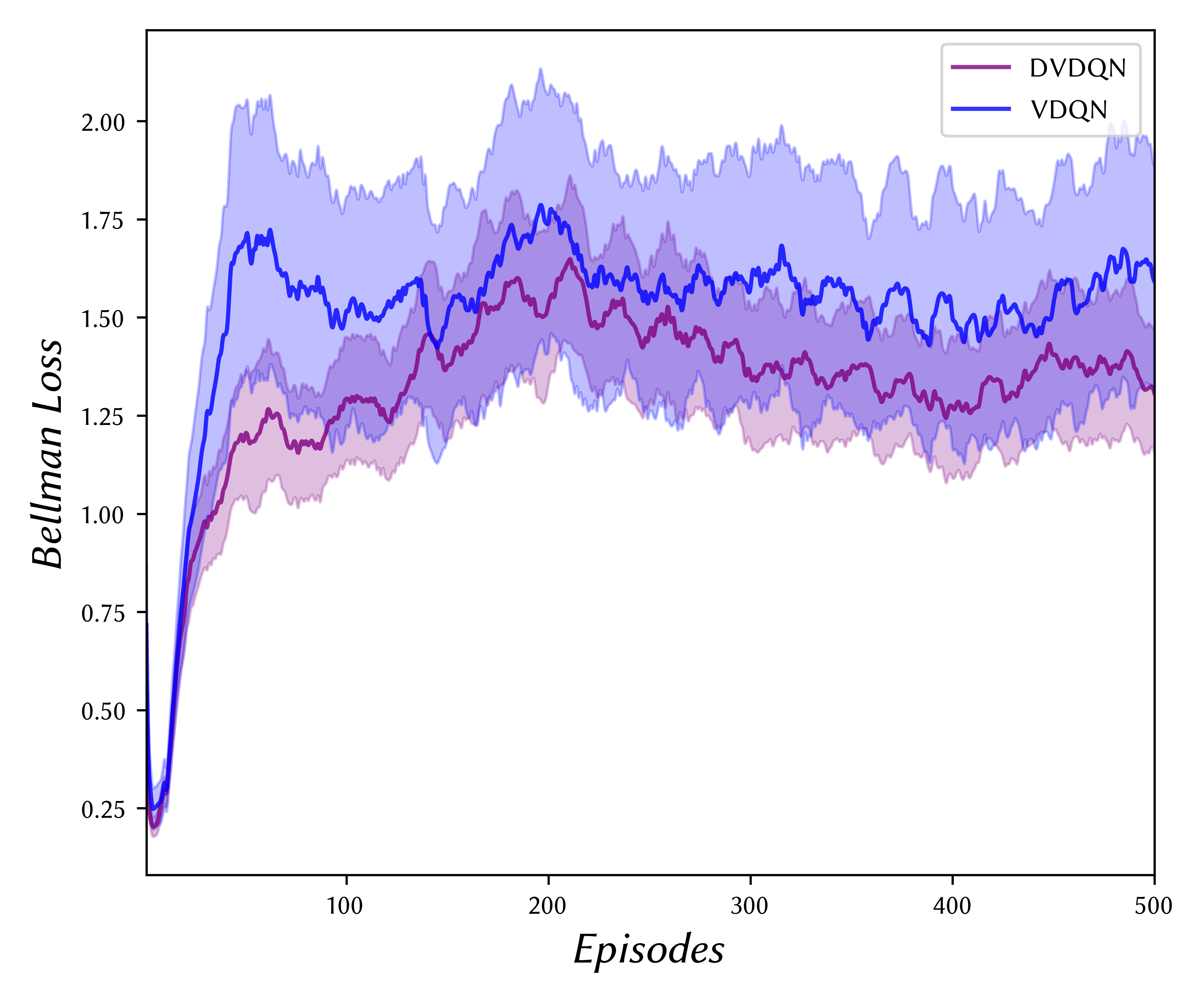}
  \vspace{-7mm}
  \caption{\texttt{Acrobot-v1} --- Bellman Error}
  \label{fig:bw-bidir-1:b}
\end{subfigure}%

\medskip

\hfill%
\begin{subfigure}{.66\textwidth}
  \centering
  \includegraphics[width=\hsize]{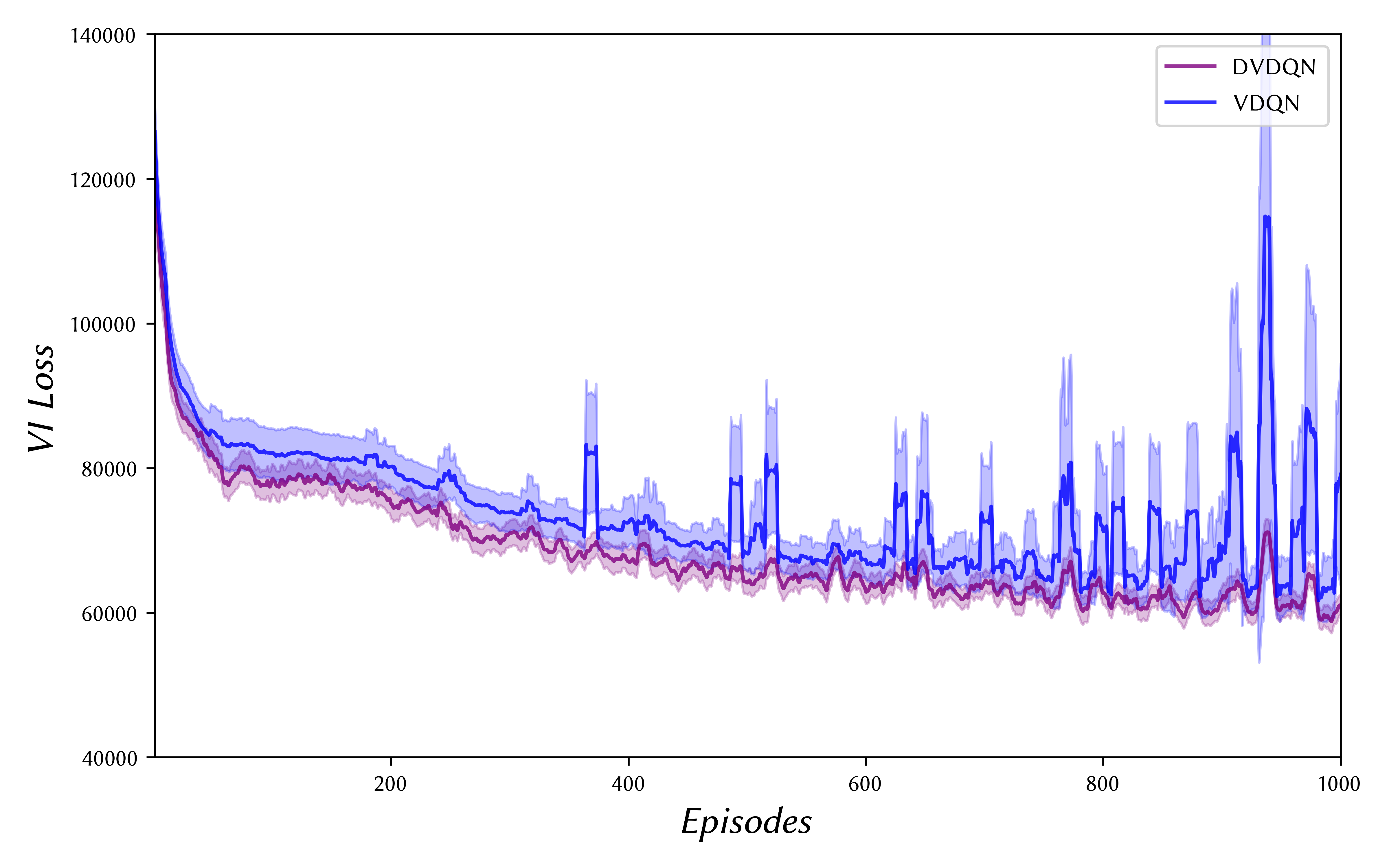}
  \vspace{-7mm}
  \caption{\texttt{MountainCar-v0} --- VI Loss}
  \label{fig:bw-bidir-1:d}
\end{subfigure}%
\hspace{0.17\textwidth}

\captionsetup{justification=centering}
\caption{Variational training losses observed during training on selected tasks. \\ Error bounds are $\pm s$, the sample std. dev.}
\label{fig:basic-vi-bellmans}
\end{figure}

\paragraph{} In these simple control tasks, DVDQN performs just as well or better than VDQN; improvements are made faster, and in some cases the final policy found is more optimal. Figure \ref{fig:basic-vi-bellmans} tells an even more interesting story --- these types of visualisation was omitted from the original VDQN paper. The most crucial improvement DVDQN appears to bring is a more stable and improved performance in VI training loss. VI loss can be conceptually thought of as the delta between the prior and approximate posterior distributions after training; decreasing VI loss implies convergence. DVDQN appears to both converge at a faster speed and exhibit a much lower variance; VDQN's peaks, representing sudden, unexpected divergence, are significantly reduced --- this was the primary goal of DVDQN.

\paragraph{} To complement the four simple control tasks in Figure \ref{fig:basic-rewards}, all four algorithms were trained against two of OpenAI Gym's Atari game environments --- \texttt{SpaceInvaders-v0} and \texttt{Pong-v0}. Only 100 episodes were run for each to try to capture how effective each algorithm's exploration strategy is in the environment.

\begin{enumerate}
    \item \texttt{SpaceInvaders-v0} --- in almost every episode DVDQN is able to accrue the greatest reward out of the four algorithms, with VDQN easily able to match the performance of the two non-variational algorithms.
    \item \texttt{Pong-v0} --- the variational algorithms appear to  fail to convert exploration to any meaningful reward, in contrast to non-variational ones which, although not close to being optimal, occasionally win one point. The VI loss curves (not presented here) appear to neither converge nor diverge; minimal progress appear to be made, which may explain the poor performance.
\end{enumerate}

\paragraph{Randomised Prior Initialisation} Osband et. al.~\cite{osb2018randomized} make an incredibly interesting argument about the unsuitability of randomised prior functions. \textit{Lemma 2} in their paper points out the following:

\begin{quote}
    Let $Y\sim \mathbb{N}(\mu_Y, \sigma_Y)$.\\
    If we train $X\sim \mathbb{N}(\mu, \sigma)$ according to mean squared error
    \begin{align*}
    \mu^{*}, \sigma^{*} \in\argmin_{\mu,\sigma} \mathbb{E} \big[ (X-Y)^2 \big] \,\,\,\text{  for X, Y independent}
    \end{align*}
    then the solution $\mu^{*} = \mu_{Y}$, $\sigma^{*} = 0$ propagates no uncertainty from Y to X.
\end{quote}

\paragraph{} This, unfortunately, does appear to encompass the prior function used by VDQNs --- the randomised uniform prior initialisation is independent to the (assumed) Gaussian distribution of the variational family. The consequences of this aren't clear, as as far as was expected the variational algorithms appear to work well. Further investigation is undoubtedly required, but this could explain the occasional failure of the inference algorithm to converge.

\subsection{Edward}
\paragraph{} One of the questions this study aimed to investigate was how impactful the use of Edward in an RL context would be. Table \ref{tab:times} presents the relative training times of each algorithm, using the number of iterations (steps taken by an actor) per second as an indicator. The training time required for these implementations of the variational algorithms are $\sim 8$x those for the non-variational. This isn't entirely surprising --- DQN/DDQN are implemented using a fairly standard Chainer design, and as such should be expected to be more heavily optimised than a manually wired Bayesian neural network. Although this may not seem entirely fair, it is important to note that beyond optimising the Edward library itself there is not much scope for improvements; a large portion of the additional running time is used to run the inference engine. Translating the VDQN/DVDQN implementations to use Pyro\footnote{\url{https://pyro.ai},~\cite{bingham2018pyro}} instead of Edward would help isolate bottlenecks, and provide a benchmark for evaluating both in the context of Deep RL. To ensure fairness, every experiment ran using 2 threads and 3 GB RAM each.

\begin{table}[]
    \centering
    \begin{tabular}{ c|c } 
     & Relative performance \\
    \hline
    DQN & 1.000 $\pm$ 0.000 \\ 
    DDQN & 0.859 $\pm$ 0.252 \\ 
    VDQN & 0.133 $\pm$ 0.029 \\ 
    DVDQN & 0.110 $\pm$ 0.031 \\
    \end{tabular}
    \vspace{5mm}
    
    \captionsetup{justification=centering}
    \caption{Relative performance of each algorithm against DQN. The error values are $\pm s$, the sample std. dev. \\ Comparisons are made using the number of iterations per second the implementation is able to achieve in each environment.}
    \label{tab:times}
\end{table}

\subsection{Avenues for Future Work}
\paragraph{} The question raised by Osband et. al. about the validity of the initial prior distribution used will be important to resolve. They present a number of interesting ideas in their paper~\cite{osb2018randomized} which could be used in lieu, particularly the introduction of $\gamma$-discounted temporal difference loss (\ref{eqn:gdtd}) --- this could be a viable replacement for Bellman error. The target and active $Q$ functions are generated from a combination of the variational families and a new prior function, $p$; a full explanation can be found in their paper.

\vspace{-4mm}
\begin{align}
\label{eqn:gdtd}
\mathcal{L}(q_{\phi}, q_{\phi^{-}}, p, D) = \sum_{t \in D} \left( r_t + \gamma \argmax_{a'}(q_{\phi^{-}} + p)(s_{t}', a') - (q_{\phi} + p)(s_{t}, a_t) \right)^2
\end{align}

\paragraph{} There are a number of improvements to that framework and additional experiments to develop understanding about VDQNs and DVDQNs further. For example, adapting the framework to accept all OpenAI Gym will open up a much wider range of evaluation contexts, such as the more advanced Atari games or Robotics simulators. This study explores a handful of the models' parameters, but a number of the VDQN/DVDQN parameters could be tuned to improve performance further.
\begin{enumerate}
    \item $\tau$ --- this implementation supports Polyak averaging~\cite{polyak} for model updates, though this study only used $\tau = 1$ (hard-copy updating).
    \item $\sigma$ --- the standard deviation of the Gaussian created for selecting the next greedy action to take.
    \item $\gamma$ --- the learning rate. Although a number were tested in this study's experiments, the behaviours of the variational and non-variational variants were very different.
    \item The number of hidden layers --- the effect of changing the size of the neural network structure is currently completely unknown. $n = 100$ layers were used here. 
\end{enumerate}


\clearpage

\clearpage

\clearpage

\clearpage

\section{Conclusion}
\paragraph{} Variational Bayesian interpretations of Deep Q Learning algorithms have proved to be promising, appearing to deliver efficient state-space exploration of complex environments. This study has provided a research-ready framework for using these approaches, as well as DVDQN, an effective optimisation technique and small novel contribution to the field. Although there are a number of interesting scenarios and promising parameter configurations left unexplored, a number of key, previously unseen, observations have been shown. These will, hopefully, provide inspiration and motivation for further exploration in this area.

\DeclareFieldFormat{titlecase}{#1}
\printbibliography

\clearpage
\appendix
\section{Reproducing the Experiments}
\label{appendix:reproduce}
\paragraph{} The evaluation framework developed alongside this report is readily available on GitHub, and has been fully tested on both macOS and Linux.
\begin{align*}
    \text{\url{https://github.com/HarriBellThomas/VDQN}}
\end{align*}

\paragraph{} The next few steps assume a fresh Ubuntu installation is being used. The process may deviate slightly if a different system is used. A helper script is included for installing the required dependencies on Linux; the process is near-identical for macOS.

\begin{quote}
    \texttt{git clone https://github.com/HarriBellThomas/VDQN.git} \\
    \texttt{cd VDQN} \\
    \texttt{./init.sh} \\
    \texttt{source env/bin/activate}
\end{quote}

\paragraph{} All Python dependencies are installed and managed inside a Python virtual environment; this helps keep the system clean, as we will be using some slightly older versions of standard libraries.

\paragraph{} There are four main files to be aware of:
\begin{enumerate}
    \item \texttt{run.py} --- this is the main entry point for running a single instance of one of the four algorithms. It accepts a number of CLI arguments for configuring the parameters it used.
    \item \texttt{DQN.py} --- this is the source file containing the implementations for both DQN and DDQN.
    \item \texttt{VDQN.py} --- this is the source file containing the implementations for both VDQN and DVDQN.
    \item \texttt{drive.py} --- this script is the driver used for running experiments at scale. It constructs a collection of 80 experiments, and iteratively loops through them.
\end{enumerate}

\paragraph{} The \texttt{run.py} script can be used as follows:

\vspace{-6mm}
\begin{align*}
    \texttt{python3 run.py } & \texttt{-{}-algorithm DQN|DDQN|VDQN|DVDQN \textbackslash{}} \\
    & \texttt{-{}-environment CartPole-v0 \textbackslash{}} \\
    & \texttt{-{}-episodes 200 \textbackslash{}} \\
    & \texttt{-{}-timesteps 200 \textbackslash{}} \\
    & \texttt{-{}-lossrate 1e-2} \\
\end{align*}

\end{document}